\documentclass[10pt,twocolumn,letterpaper]{article}
\usepackage{titling}
\usepackage[pagenumbers]{cvpr} 

\usepackage{graphicx}
\usepackage{array}
\usepackage{amsmath}
\usepackage{amssymb}
\usepackage{booktabs}
\usepackage{multirow}
\usepackage{multirow}
\usepackage[table]{xcolor}
\usepackage{marvosym}
\definecolor{mygray}{gray}{.92}
\usepackage[pagebackref=true,breaklinks=true,linkcolor=red,anchorcolor=blue, citecolor=green,colorlinks,bookmarks=false,backref=false]{hyperref}
\usepackage[capitalize]{cleveref}
\crefname{section}{Sec.}{Secs.}
\Crefname{section}{Section}{Sections}
\Crefname{table}{Table}{Tables}
\crefname{table}{Tab.}{Tabs.}

\def\cvprPaperID{6230} 
\def\confName{CVPR}
\def\confYear{2023}

\makeatletter
\def\thanks#1{\protected@xdef\@thanks{\@thanks
		\protect\footnotetext{#1}}}
\makeatother

\begin{document}
	
	\title{Mapping Degeneration Meets Label Evolution: Learning Infrared Small Target Detection with Single Point Supervision}
	
	\author{Xinyi~Ying$^{1}$, Li~Liu$^{1}$, Yingqian~Wang$^{1}$, Ruojing~Li$^{1}$, Nuo~Chen$^{1}$, Zaiping~Lin$^{1}$\textsuperscript{\Letter},\\ Weidong~Sheng$^{1}$, Shilin~Zhou$^{1}$\\
		$^{1}$National University of Defense Technology\\
		\{yingxinyi18, wangyingqian16, liruojing, chennuo97, linzaiping, slzhou\}@nudt.edu.cn, \\
		dreamliu2010@gmail.com, shengweidong1111@sohu.com		
		\thanks{This work was supported by National Key Research and Development Program of China No. 2021YFB3100800.}}
	\date{}
	\maketitle
	
	
	\begin{abstract}
		Training a convolutional neural network (CNN) to detect infrared small targets in a fully supervised manner has gained remarkable research interests in recent years, but is highly labor expensive since a large number of per-pixel annotations are required.
		To handle this problem, in this paper, we make the first attempt to achieve infrared small target detection with point-level supervision.
		Interestingly, during the training phase supervised by point labels, we discover that CNNs first learn to segment a cluster of pixels near the targets, and then gradually converge to predict groundtruth point labels. 
		Motivated by this ``mapping degeneration'' phenomenon, we propose a label evolution framework named label evolution with single point supervision (LESPS) to progressively expand the point label by leveraging the intermediate predictions of CNNs. In this way, the network predictions can finally approximate the updated pseudo labels, and a pixel-level target mask can be obtained to train CNNs in an end-to-end manner.
		We conduct extensive experiments with insightful visualizations to validate the effectiveness of our method. Experimental results show that CNNs equipped with LESPS can well recover the target masks from corresponding point labels, {and can achieve over 70\% and 95\% of their fully supervised performance in terms of pixel-level intersection over union ($IoU$) and object-level probability of detection ($P_d$), respectively. Code is available at \href{https://github.com/XinyiYing/LESPS}{https://github.com/XinyiYing/LESPS}.}
	\end{abstract}
	
	\vspace{-0.3cm}
	\section{Introduction}
	\label{sec:intro}
	
	\begin{figure}[t]
		\centering\includegraphics[width=8cm]{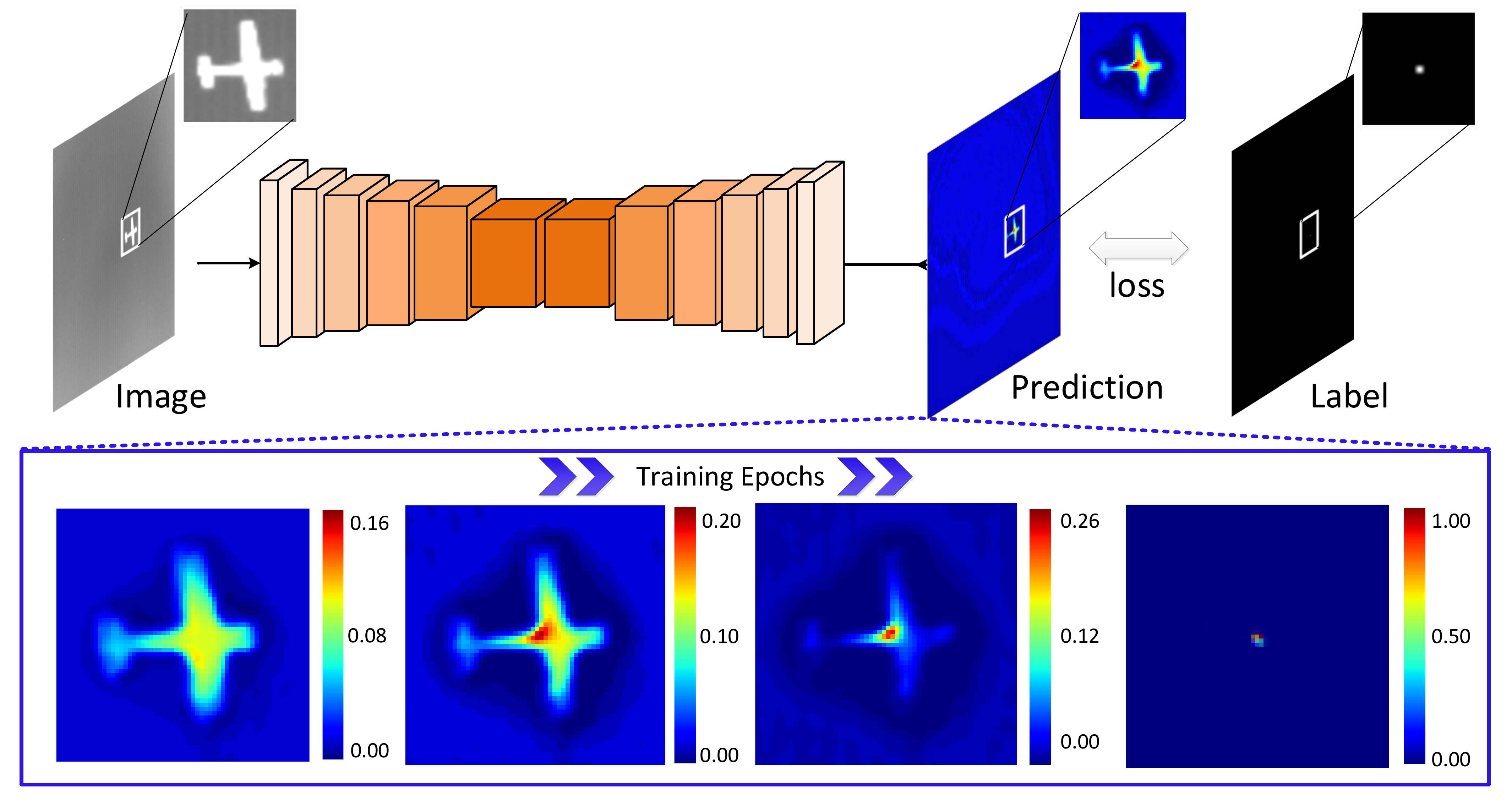}
		\vspace{-0.1cm}
		\caption{{An illustration of mapping degeneration under point supervision. CNNs always tend to segment a cluster of pixels near the targets with low confidence at the early stage, and then gradually learn to predict GT point labels with high confidence.}}\label{fig:nonlinear1}
		\vspace{-0.45cm}
	\end{figure}
	
	Infrared small target detection has been a longstanding, fundamental yet challenging task in infrared search and tracking systems, and has various important applications in civil and military fields \cite{survey1,survey2}, including traffic monitoring \cite{ISNet,DNA-Net}, maritime rescue \cite{Mingjin1,Mingjin2} and military surveillance \cite{ACM,MoCoPnet}.
	Due to the rapid response and robustness to fast-moving scenes, single-frame infrared small target (SIRST) detection methods have always attracted much more attention, and numerous methods have been proposed. Early methods, including filtering-based \cite{tophat,Max-Median}, local contrast-based \cite{LCM,TLLCM} and low rank-based \cite{IPI,MSLSTIPT} methods, require complex handcrafted features with carefully tuned hyper-parameters. Recently, compact deep learning has been introduced in solving the problem of SIRST detection \cite{MDvsFA,DNA-Net,ISNet}. However, there are only a few attempts, and its potential remains locked, unlike the extensive explorations of deep learning for natural images. This is mainly due to potential reasons, including lack of large-scale, accurately annotated datasets and high stake application scenarios.
	
	Infrared small targets are usually of very small size, weak, shapeless and textureless, and are easily submerged in diverse complex background clutters. As a result, directly adopting existing popular generic object detectors like RCNN series {\cite{RCNN1,RCNN2,RCNN3,RCNN0}}, YOLO series {\cite{YOLO1,YOLO2,YOLO3}} and SSD {\cite{SSD1}} to SIRST detection cannot produce satisfactory performance. Realizing this, researchers have been focusing on developing deep networks tailored for infrared small targets by adequately utilizing the domain knowledge. However, most existing deep methods for SIRST detection \textcolor{black}{\cite{DNA-Net,ISNet,ALCNet}} are fully supervised, which usually requires a large dataset with accurate target mask annotations for training. Clearly, this is costly \cite{PointSeg1,PointSeg2}.
	
	Therefore, a natural question arises: \textit{Can we develop a new framework for SIRST detection with single point supervision}? In fact, to substantially reduce the annotation cost for object detection tasks, weakly supervised object detection methods with point supervision \cite{PointOB1,PointOB2,PointSeg1,PointSeg2} have been studied in the field of computer vision. Although these weakly supervised methods achieve promising results, they are not designed for the problem of SIRST detection, and \textcolor{black}{the class-agnostic labels (\textit{i.e.,} only foreground and background) of infrared small targets hinder their applications \cite{CAM1,CAM2}.} Therefore, in this work, we intend to conduct the first study of weakly supervised SIRST detection with single-point supervision.
	
	{A key motivation of} this work comes from an interesting {observation} during the training of SIRST detection networks. That is, with single point labels serving as supervision, CNNs always tend to segment a cluster of pixels near the targets with low confidence at the early stage, and then gradually learn to predict groundtruth (GT) point labels with high confidence, as shown in Fig.~\ref{fig:nonlinear1}. 
	It reveals the fact that {region-to-region} mapping is the intermediate result of the final region-to-point mapping\footnote{{ ``region-to-region mapping” represents the mapping learned by CNNs from target regions in images to a cluster of pixels near the targets, while ``region-to-point mapping” represents the mapping from target regions in images to the GT point labels.}}. {We attribute this ``mapping degeneration'' phenomenon to the special imaging mechanism of infrared system \cite{DNA-Net,ISNet}, the local contrast prior of infrared small targets \cite{LCM,ALCNet}, and the easy-to-hard learning property of CNNs \cite{Deep_prior}, in which the first two factors result in extended mapping regions beyond the point labels, and the last factor contributes to the degeneration process.} 
	
	{Based on the aforementioned discussion, in this work, we propose a novel framework for the problem of weakly supervised SIRST detection, dubbed label evolution with single point supervision (LESPS).} {Specifically, LESPS leverages the intermediate network predictions in the training phase to update the current labels, which serve as supervision until the next label update. Through iterative label update and network training, the network predictions can finally approximate the updated pseudo mask labels, and the network can be simultaneously trained to achieve pixel-level SIRST detection in an end-to-end\footnote{Different from generic object detection \cite{End1,End2}, ``end-to-end'' here represents achieving point-to-mask label regression and direct pixel-level inference in once training.} manner.}
	
	{Our main contributions are summarized as: (1) We present the first study of weakly supervised SIRST detection, and introduce LESPS that can significantly reduce the annotation cost.} (2) {We discover the mapping degeneration phenomenon, and leverage this phenomenon to automatically regress pixel-level pseudo labels from the given point labels via LESPS.} (3) Experimental results show that our framework can be applied to different existing SIRST detection networks, and enable them to {achieve over 70\% and 95\% of its fully supervised performance in terms of pixel-level intersection over union ($IoU$) and object-level probability of detection ($P_d$), respectively.} 
	
	\section{Related Work}
	\label{sec:related}
	
	\textbf{SIRST Detection.} 
	In the past decades, various methods have been proposed, including early traditional paradigms (\textit{e.g., }filtering-based methods \cite{tophat,Max-Median}, local contrast-based methods \cite{LCM,RLCM,MSLCM,MSPCM,WSLCM,TLLCM}, low rank-based methods \cite{IPI,RIPT,NRAM,low_rank1,PSTNN,MSLSTIPT}) and recent deep learning paradigms \cite{MDvsFA,ACM,ALCNet,DNA-Net,ISNet,Mingjin1,Mingjin2,ISTDet,RISTDnet}. 
	Compared to traditional methods, which require delicately designed models and carefully tuned hyper-parameters, convolutional neural networks (CNNs) can learn the non-linear mapping between input images and GT labels in a data-driven manner, and thus generalize better to real complex scenes. As the pioneering work, Wang \textit{et al.} \cite{MDvsFA} first employed a generative adversarial network to achieve a better trade-off between miss detection and false alarm. Recently, more works focus on customized solutions of infrared small target. Specifically, Dai \textit{et al.} \cite{ACM} specialized an asymmetric contextual module, and further incorporated local contrast measure \cite{ALCNet} to improve the target contrast. Li \textit{et al.} \cite{DNA-Net} preserved target information by repetitive feature fusion. Zhang \textit{et al.} \cite{ISNet} aggregated edge information to achieve shape-aware SIRST detection. Zhang \textit{et al.} \cite{Mingjin1,Mingjin2} explored cross-level correlation and transformer-based method \cite{Transformer} to predict accurate target mask. Wu \textit{et al.} \cite{UIU-Net} customized a UIU-Net framework for multi-level and multi-scale feature aggregation.
	In conclusion, existing works generally focus on compact architectural designs to pursue superior performance in a fully supervised manner. However, due to the lack of a large number of public datasets \cite{MDvsFA,ACM,DNA-Net} with per-pixel annotations, the performance and generalization of CNNs are limited. In addition, per-pixel manual annotations are time-consuming and labor-intensive. Therefore, we focus on achieving good pixel-level SIRST detection with weaker supervision and cheaper annotations.%
	
	{\textbf{Weakly Supervised Segmentation with Points.}}
	Recently, point-level annotation has raised more attention in dense prediction tasks such as object detection \cite{PointOB1,PointOB2,PointOB3}, crowd counting \cite{PointCount1,PointCount2,PointCount3,PointCount4} and image segmentation \cite{PointSeg1,PointSeg2,PointSeg3,PointSeg4,PointSeg5,PointSeg6,PointSeg7}. We mainly focus on image segmentation in this paper. Specifically, Bearman \textit{et al.} \cite{PointSeg7} made the first attempt to introduce an objectiveness potential into a pointly supervised training loss function to boost segmentation performance. Qian \textit{et al.} \cite{PointSeg5} leveraged semantic information of several labeled points by a distance metric loss to achieve scene parsing. Zhang \textit{et al.} \cite{PointSeg4} proposed an inside-outside guidance approach to achieve instance segmentation by five elaborate clicks. Cheng \textit{et al.} \cite{PointSeg1} designed to provide ten randomly sampled binary point annotations within box annotations for instance segmentation. Li \textit{et al.} \cite{PointSeg2} encoded each instance with kernel generator for panoptic segmentation to achieve 82\% of fully-supervised performance with only twenty randomly annotated points. In contrast to these approaches employing complicated prior constraints to segment large generic objects with rich color and fine textures by several elaborate points, we fully exploit the local contrast prior of infrared small target to progressively evolve pseudo masks by single coarse point without any auxiliaries in an end-to-end manner. 
	
	\section{The Mapping Degeneration Phenomenon} 
	
	\label{sec:nonlinear}
	In this section, we first describe the mapping degeneration phenomenon together with our intuitive explanation. Then we conduct experiments under single-sample and many-sample training schemes to demonstrate the generality of degeneration, and investigate the influence of generalization on degeneration.
	
	As shown in Fig.~\ref{fig:nonlinear1}, given an input image and the corresponding GT point label, we employ U-Net \cite{U-Net} as the baseline SIRST detection network for training.  
	It can be observed that, in the early training phase, network predicts a cluster of pixels near the targets with low confidence. As training continues, the network prediction finally approximates GT point label with gradually increased confidence. We name this phenomenon as ``mapping degeneration'', and attribute the following reasons to this phenomenon.
	1) \textit{Special imaging mechanism of infrared systems} \cite{DNA-Net,ISNet}: Targets only have intensity information without structure and texture details, resulting in highly similar pixels within the target region.
	2) \textit{High local contrast of infrared small targets} \cite{LCM,ALCNet}: Pixels within the target region are much brighter or darker with high contrast against the local background clutter.
	3) \textit{Easy-to-hard learning property of CNNs} \cite{Deep_prior}: CNNs always tend to learn simple mappings first, and then converge to difficult ones. Compared with region-to-point mapping, region-to-region mapping is easier, and thus tends to be the intermediate result of region-to-point mapping.
	In conclusion, the unique characteristics of infrared small targets result in extended mapping regions beyond point labels, and CNNs contribute to the mapping degeneration process.
	
\begin{figure}[t]
		\centering\includegraphics[width=8.5cm]{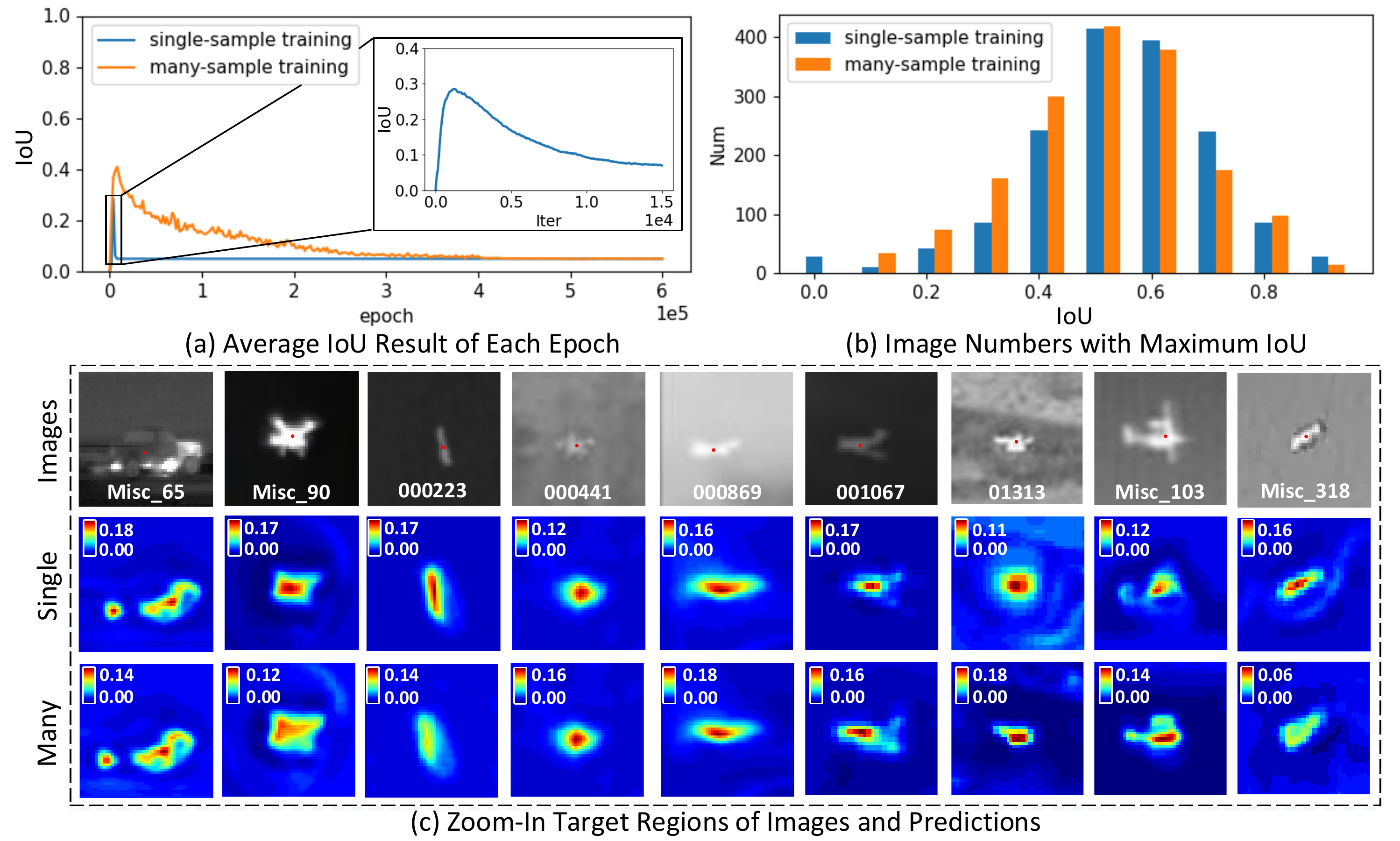}
		\vspace{-0.6cm}
		\caption{{Quantitative and qualitative illustrations of mapping degeneration in CNNs. 
		}}\label{fig:mpvsg}
		\vspace{-0.55cm}
	\end{figure}

	It is worth noting that the mapping degeneration phenomenon is a general phenomenon in various scenes with infrared small targets. Specifically, we use the training datasets (including 1676 images and their corresponding centroid point label, see details in Section~\ref{sec:imple}) to train U-Net under a single-sample training scheme (\textit{i.e.,} training one CNN on each image). For quantitative analyses, we employ the $IoU$ results between positive pixels in predictions (\textit{i.e.,} pixels with confidence higher than half of its maximum value) and GT mask label. Average $IoU$ results of 1676 CNNs at each epoch are shown by the blue curve in Fig.~\ref{fig:mpvsg}(a), while the number of training samples with maximum $IoU$ during training phase falling in a threshold range of $[i,i+0.1], (i=0,0.1,\cdots,0.9)$ is illustrated via blue bars in Fig.~\ref{fig:mpvsg}(b). It can be observed from the zoom-in curve and bars that mapping degeneration is a general phenomenon with point supervision, and U-Net can achieve $IoU>0.5$ on more than 60\% of the training images.
	
	In addition, we conduct experiments to train U-Net under a many-sample training scheme (\textit{i.e.,} training one CNN using all images which contain abundant targets with various sizes and shapes) to investigate the effect of generalization on mapping degeneration. Average $IoU$ results of 1676 images are shown by orange curve in Fig.~\ref{fig:mpvsg}(a). It can be observed that many-sample training scheme needs more time to converge. Moreover, Fig.~\ref{fig:mpvsg}(b) shows that orange bars are slightly lower than blue ones on larger $IoU$ values (\textit{i.e.,} 0.5-1.0). It is demonstrated that generalization decelerates but aggravates mapping degeneration. Figure~\ref{fig:mpvsg}(c) shows some zoom-in target regions of images and their predictions under these two training schemes. It can be observed that CNNs can effectively segment a cluster of target pixels under both training schemes in a size-aware manner. 
	
	Therefore, an intuitive assumption arises: Can we leverage the intermediate results of CNNs to regress masks? A simple early stopping strategy seems to be a positive answer but is indeed unpractical since mapping degeneration is influenced by various factors, including target intensity, size, shape, and local background clutter (see details in Section~\ref{exp:MP}). Consequently, there is no fixed optimal stopping epoch for all situations. These observations motivate us to design a special label evolution framework to well leverage the mapping degeneration for pseudo mask regression.
	
		
\begin{figure}[t]
	\vspace{-0.4cm}
	\centering\includegraphics[width=8.5cm]{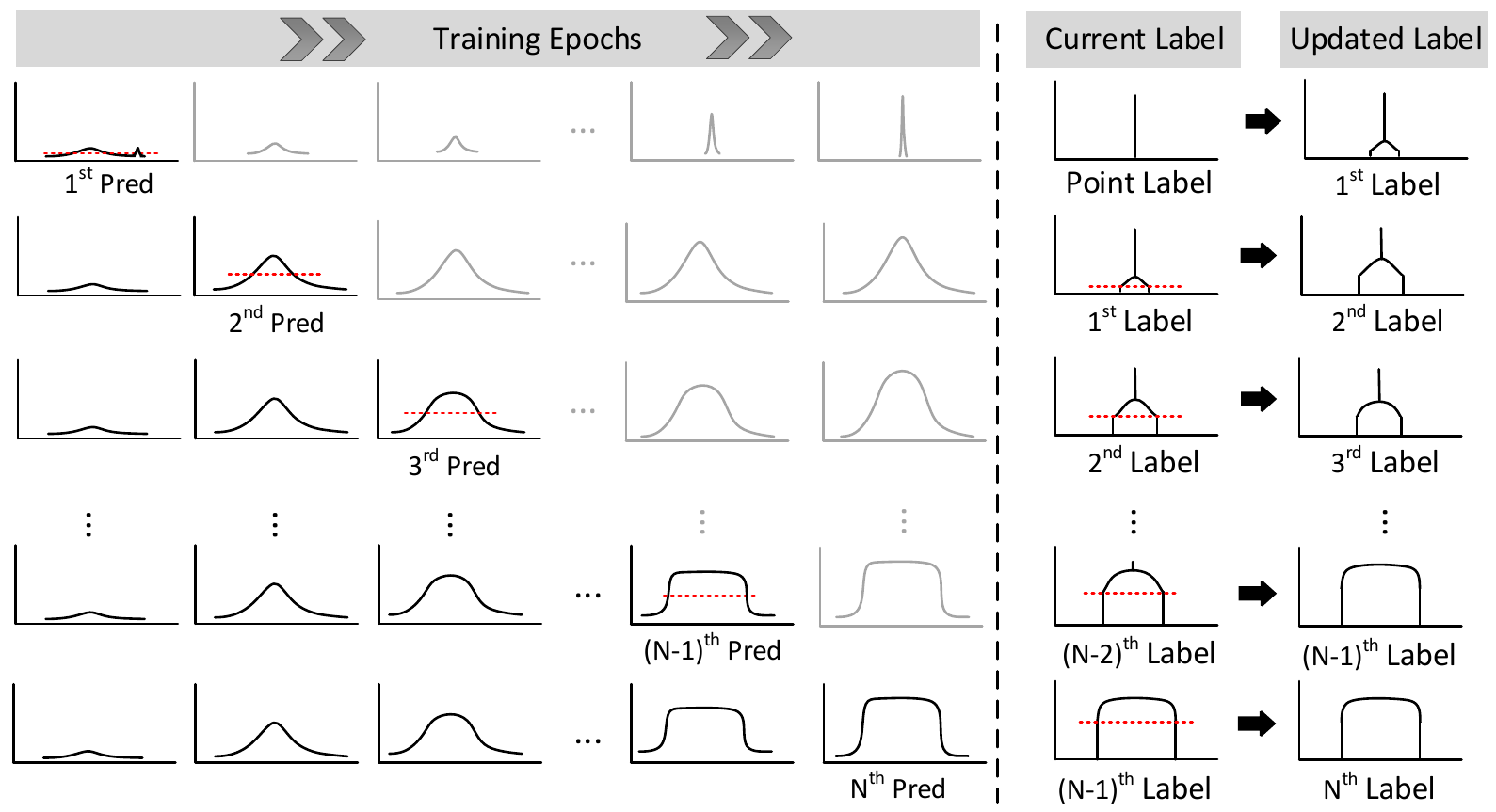}
	\vspace{-0.6cm}
	\caption{{An illustration of  label evolution with single point supervision (LESPS). During training, intermediate predictions of CNNs are used to progressively expand point labels to mask labels. Black arrows represent each round of label updates.}}\label{fig:method}
	\vspace{-0.4cm}
\end{figure}

	\section{The Label Evolution Framework}
	Motivated by mapping degeneration, {we propose a label evolution framework named label evolution with single point supervision (LESPS) to leverage the intermediate network predictions in the training phase to update labels.} As training continues, the network predictions approximate the updated pseudo mask labels, and network can simultaneously learn to achieve pixel-level SIRST detection in an end-to-end manner. Here, we employ a toy example of 1D curves for easy understanding. As shown in Fig.~\ref{fig:method}, sub-figures on the left of the dotted line represent the network predictions. Note that, the black curves denote the intermediate predictions within LESPS, while the gray curves represent virtual results produced by the network without label update. On the right of the dotted line, the first and second columns of sub-figures represent current labels and updated labels, respectively, and black arrows represent each round of label update. The overall framework can be summarized as follows. With point label serving as supervision, in the $1^{st}$ round label update after initial training, the predictions are used to update the current point label to generate the $1^{st}$ updated label, which is then used to supervise the network training until the $2^{nd}$ round label update. Through iterative label updates and network training, CNNs can incorporate the local contrast prior to gradually recover the mask labels. From another viewpoint, label evolution consistently updates the supervision to prevent mapping degeneration, and promotes CNNs to converge to the easy region-to-region mapping.

\begin{figure}[t]
		\centering
		\vspace{-0.4cm}
		\includegraphics[width=8.2cm]{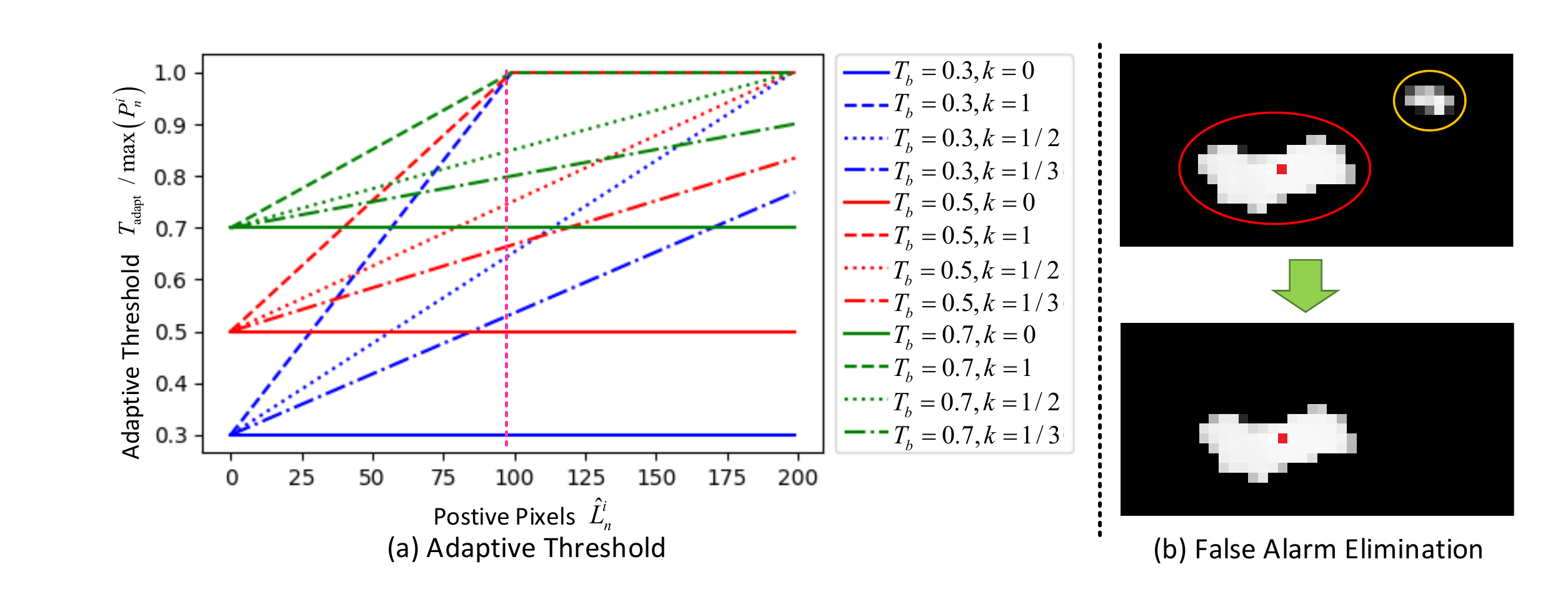}
		\vspace{-0.1cm}
		\caption{(a) Adaptive threshold $T_{adapt}$ with respect to positive pixels $\hat{L}_n^{i}$ and hyper-parameters $k$, $T_b$. Pink dotted line represents the constant $hwr$. (b) An illustration of false alarm elimination. Red circle and dot represent positive pixels and centroid point of label. Orange circle represents false alarms.} \label{thre}
		\vspace{-0.35cm}
	\end{figure}

	Taking the $n^{th}$ update as an example, given the current label $L_{n}$ and the network prediction $P_{n}$, we perform label update for each target, which consists of three steps: candidate pixel extraction, false alarm elimination, and weighted summation between candidate pixels and current labels. Specifically, the $d\times d$ local neighborhoods of the $i^{th}$ target in label $L_{n}$ and prediction $P_{n}$ are cropped based on the centroid of the positive pixels\footnote{The value of a pixel is higher than 0.5, which represents that the pixel is more likely to be positive than negative \cite{DNA-Net, ALCNet}} in label (\textit{i.e.,} $\hat{L}_{n}$). Then to reduce error accumulation for label update (see Section~\ref{sec:exp-thresh} for details), we employ an adaptive threshold (the red dotted line in Fig.~\ref{fig:method}) to extract the local neighborhood candidate pixels (predictions higher than the red dotted line in Fig.~\ref{fig:method}). The process can be defined as:
	\begin{align}\label{eq-seg}
		\abovedisplayshortskip=0pt
		\belowdisplayshortskip=0pt
		\abovedisplayskip=0pt
		\belowdisplayskip=0pt
		C_{n}^{i} &= P_{n}^{i}\odot (P_{n}^{i}>T_{adapt}),
	\end{align}
	where $C_{n}^{i}$ is the candidate pixels, and $\odot$ represents element-wise multiplication. $T_{adapt}$ is the adaptive threshold that correlated to the current prediction $P_{n}^{i}$ and the positive pixels in label $\hat{L}_{n}^{i}$, and can be calculated according to:
	\begin{align}\label{eq-thresh}
		T_{adapt}&= max(P_{n}^{i})(T_{b}+k(1-T_{b})\hat{L}_{n}^{i}/(hwr)),
	\end{align}
	where $h$, $w$ are the height and width of input images, and $r$ is set to 0.15\% \cite{DNA-Net,ACM}. As shown in Fig.~\ref{thre} (a), $T_b$ is the minimum threshold, and $k$ controls the threshold growth rate. An increasing number of $\hat{L}_{n}^{i}$ leads to the increase of the threshold, which can reduce error accumulation of low contrast targets and strong background clutter.
			
	\begin{figure*}[t]
		\vspace{-0.3cm}
		\centering\includegraphics[width=17.5cm]{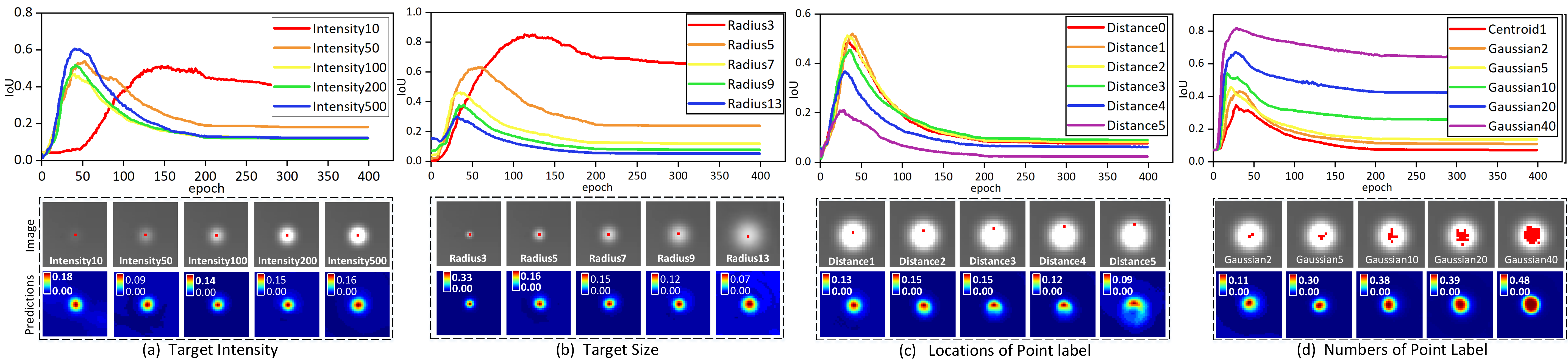}
		\vspace{-0.6cm}
		\caption{$IoU$ and visualize results of mapping degeneration with respect to different characteristics of targets (\textit{i.e.,}(a) intensity, (b) size,) and point labels (\textit{i.e.,}(c) locations and (d) numbers). We visualize the zoom-in target regions of input images with GT point labels (\textit{i.e.,} red dots in images) and corresponding CNN predictions (in the epoch reaching maximum $IoU$).}\label{fig:mp_Quan}
		\vspace{-0.3cm}
	\end{figure*}
	
	To eliminate false alarms by local neighborhood noise, we exclude the eight connective regions of candidate pixels that have no intersection with positive pixels of labels, as shown in Fig.~\ref{thre} (b). This process is defined as:
	\begin{align}\label{eq-eight}
		E_{n}^{i}=C_{n}^{i}\odot F_{n}^{i},
	\end{align}
	where $E_{n}^{i}$ is the candidate pixels after false alarm elimination, and $F_{n}^{i}$ is the mask against false alarm pixels. 
	
	We then perform average weighted summation between candidate pixels $E_{n}^{i}$ and current label $L_{n}^{i}$ to achieve label update. The process can be formulated as:
	\begin{align}\label{eq-update}
		L_{n+1}^{i} &= L_{n}^{i}\odot (1-N_{n}^{i})+  \frac{L_{n}^{i}+E_{n}^{i}}{2}\odot N_{n}^{i},
	\end{align}
	where $L_{n+1}^{i}$ is the updated label in the ${n}^{th}$ round, which serves as new supervision for training in the ${n+1}^{th}$ round, and $N_{n}^{i}=(P_{n}^{i}>T_{adapt})\odot F_{n}^{i}$. Note that, the first term represents GT labels below red dotted lines, and the second term represents the average weighted summation between predictions and GT labels above red dotted lines.

	It is worth noting that we provide three conditions to ensure network convergence: 1) Average weighted summation between predictions and labels promotes CNNs to converge as predictions approximate labels. 2) Pixel-adaptive threshold increases with the increase of positive pixels in updated labels, which slows down or suspends the label update. 3) As label evolution introduces more target information for training, CNNs grow to mature, and learn to distinguish targets from backgrounds.
	
	To determine the exact epoch number to start label evolution, we employ a network independent loss-based threshold $T_{loss}$ instead of a network dependent epoch-based threshold to start label evolution. That is, we start the label evolution when the loss of the last epoch is lower than $T_{loss}$, and then perform label update every $f$ epochs until the end of training. Note that, our loss-based threshold is a coarse threshold to ensure that networks attend to targets instead of background clutter.
	
	
	\section{Experiments}
	\label{sec:Experiments}
	In this section, we first describe the implementation details, and then make comprehensive analyses of the mapping degeneration phenomenon and our label evolution framework. In addition, we apply our method to the state-of-the-art SIRST detection methods with point supervision, and make comparisons with their fully supervised counterparts. Moreover, we make comparisons of our dynamic updated pseudo labels with fixed pseudo labels, and discuss the calculation of loss function.
			
	\begin{figure*}[t]
		\vspace{-0.3cm}
		\centering\includegraphics[width=17.5cm]{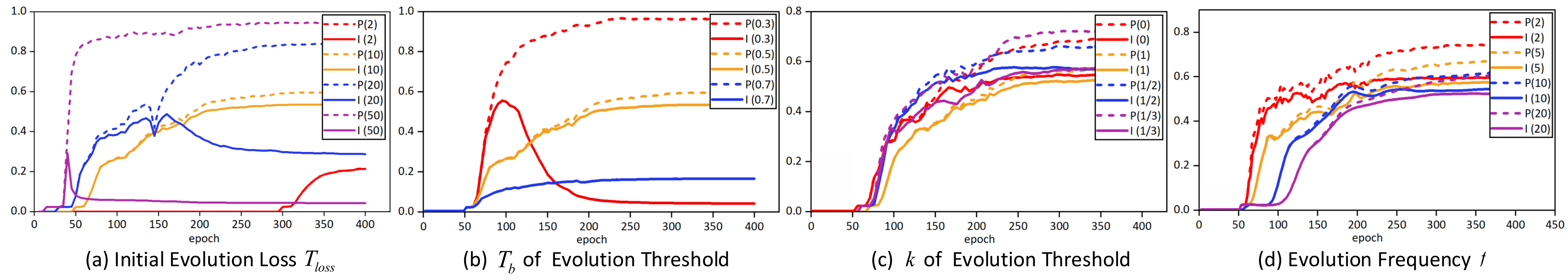}
		\vspace{-0.5cm}
		\caption{$PA$ (P) and $IoU$ (I) results of LESPS with respect to (a) initial evolution loss $T_{loss}$, (b) $T_{b}$ and (c) $k$ of evolution threshold, and (d) evolution frequency $f$.}\label{fig:LB}
		\vspace{-0.1cm}
	\end{figure*}
	
	\subsection{Implementation Details}\label{sec:imple}
	Three public datasets SIRST \cite{ALCNet}, NUDT-SIRST \cite{DNA-Net}, and IRSTD-1K \cite{ISNet} are used in our experiments. 
	We followed \cite{DNA-Net} to split the training and test sets of SIRST and NUDT-SIRST, and followed \cite{ISNet} to split IRSTD-1K. We employed two pixel-level metrics (\textit{i.e., } intersection over union ($IoU$) and pixel accuracy ($PA$)) and two target-level metrics (\textit{i.e., }probability of detection ($P_d$) and false-alarm rate ($F_a$)) for performance evaluation.
	
	During training, all images were normalized and randomly cropped into patches of size 256$\times$256 as network inputs. 
	We augmented the training data by random flipping and rotation. 
	Due to the extreme positive-negative sample imbalance (less than 10 vs. more than 256$\times$256)  in SIRST detection with point supervision, we employed focal loss\footnote{Focal loss is calculated between predition and current evolved labels to supervise the network training until the next round label update.} \cite{focal_loss} to stabilize the training process. 
	All the networks were optimized by the Adam method \cite{Adam}. 
	Batch size was set to 16, and learning rate was initially set to 5$\times$10$^{-4}$ and reduced by ten times at the 200$^{th}$ and 300$^{th}$ epochs. We stopped training after 400 epochs.
	All models were implemented in PyTorch \cite{Pytorch} on a PC with an Nvidia GeForce 3090 GPU.
	\subsection{Model Analyses} \label{Exp:Degeneration}
	
	\subsubsection{Analyses of Mapping Degeneration}\label{exp:MP}
	
	We use synthetic images (simulated targets and real backgrounds \cite{DNA-Net}) to investigate the mapping degeneration phenomenon with respect to different characteristics of targets (\textit{i.e.,} intensity and size\footnote{Shape, and local background clutter are investigated in supplemental material.}) and point labels (\textit{i.e.,} locations and numbers).
	We employ U-Net \cite{U-Net} as the baseline detection network, and use centroid point as GT label if not specified. We calculate $IoU$ between positive pixels in predictions and GT mask labels of each epoch for quantitative analyses. In addition, we visualize the zoom-in target regions of simulated images with GT point labels (\textit{i.e.,} red dots) and corresponding CNN predictions (in the epoch reaching maximum $IoU$). To reduce training randomness, we show the average $IoU$ results and visualization results over 100 runs.
	
	\textbf{Target Intensity.} We simulate Gaussian-based extended targets with different peak values (\textit{i.e.,} 10, 50, 100, 200, 500)  to investigate the influence of target intensity on mapping degeneration. Quantitative results in Fig.~\ref{fig:mp_Quan}(a) show that intensity higher than 100 leads to a positive correlation between intensity and maximum $IoU$, while lower intensity leads to a negative one. 
	In addition, curve ``intensity10'' reaches maximum $IoU$ at around epoch 150 while others are less than 50, which demonstrates that over-small intensity decelerates degeneration. 
	Visualization results show that our method can well highlight target regions under various intensities.

	\textbf{Target Size.} We simulate Gaussian-based extended targets with different radii (\textit{i.e.,} 3, 5, 7, 9, 13) to investigate the influence of target size on mapping degeneration. Quantitative results in Fig.~\ref{fig:mp_Quan}(b) show that larger target size leads to lower maximum $IoU$ and less time to reach maximum $IoU$. That is because, size discrepancy between targets and GT point labels increases as target size increases, which aggravates and accelerates mapping degeneration. 
	Visualization results show that CNNs can predict a cluster of pixels in a size-aware manner, and the peak values of predictions decrease as  target size increases. 
	
	
	
	\textbf{Locations of Point Label.} 
	We simulate a Gaussian-based extended target (with intensity 500 \& radius 13), and place point labels at different distances away from the target centroid to investigate the influence of point label locations on mapping degeneration. Results in Fig.~\ref{fig:mp_Quan}(c) show that small perturbations of label locations (less than 3 pixels) have a minor influence on the maximum $IoU$ results. However, severe location perturbations (larger than 3 pixels) can lead to a significant maximum $IoU$ drop, and the drop is more obvious when the point label is close to the edge. Note that, the same targets with different label locations reach maximum $IoU$ at the same time, which demonstrates that the speed of mapping degeneration is irrelevant to the position of labels.
		
	\begin{table}
		\centering
		\vspace{-0.2cm}
		\caption{Average $IoU$ ($\times 10^2$), $P_d$ ($\times 10^2$) and $F_a$($\times 10^6$) values on {SIRST\cite{ALCNet}} {NUDT-SIRST} \cite{DNA-Net} and {IRSTD-1K} \cite{ISNet} achieved by DNA-Net with (w/) and without (w/o) LESPS under centroid, coarse point supervision together with full supervision.}\label{Tab:effect}
		\scriptsize
		\vspace{-0.1cm}
		\setlength{\tabcolsep}{1.0mm}
		{\begin{tabular}{|l|ccc|ccc|ccc|}
				\hline
				\multirow{2}*{Method}&\multicolumn{3}{c|}{Centroid}&\multicolumn{3}{c|}{Coarse}&\multicolumn{3}{c|}{Full}\\\cline{2-10}
				&$IoU$&$P_d$&$F_a$&$IoU$&$P_d$&$F_a$&$IoU$&$P_d$&$F_a$\\\hline
				w/o LESPS&5.12&89.19&0.68&2.96&49.89&0.30 &\multirow{2}*{75.67}&\multirow{2}*{96.18}&\multirow{2}*{22.94}\\\cline{1-7}
				w/ LESPS&57.34 &91.87&20.24&56.18&91.49&18.32&&&\\\hline
		\end{tabular}}
		\vspace{-0.3cm}
	\end{table}
	
	
	\textbf{Number of Points in Label.}
	We simulate a Gaussian-based extended target (with intensity 500 \& radius 13) and set different numbers of points in labels to investigate its influence on mapping degeneration. Quantitative results in Fig.~\ref{fig:mp_Quan}(d) show that as the number of points increases, CNNs can learn more target information to achieve higher maximum $IoU$ results. In addition, the speed of mapping degeneration is irrelevant to the point number.
	Visualization results show that peak values of predictions increase as the number of points increases, which demonstrates that stronger supervision alleviates mapping degeneration. The conclusion well supports our label evolution framework.

	\subsubsection{Analyses of the Label Evolution Framework} \label{Exp:LESPS}
	
	In this subsection, we conduct experiments to investigate the effectiveness and the optimal parameter settings of our label evolution framework (\textit{i.e.,} LESPS). 
	We employ $PA$ and $IoU$ between the positive pixels in updated labels and the GT mask labels to quantitatively evaluate the accuracy and expansion degree of the current label. 
			
	\begin{figure}[t]
		\vspace{-0.4cm}
		\centering\includegraphics[width=8.5cm]{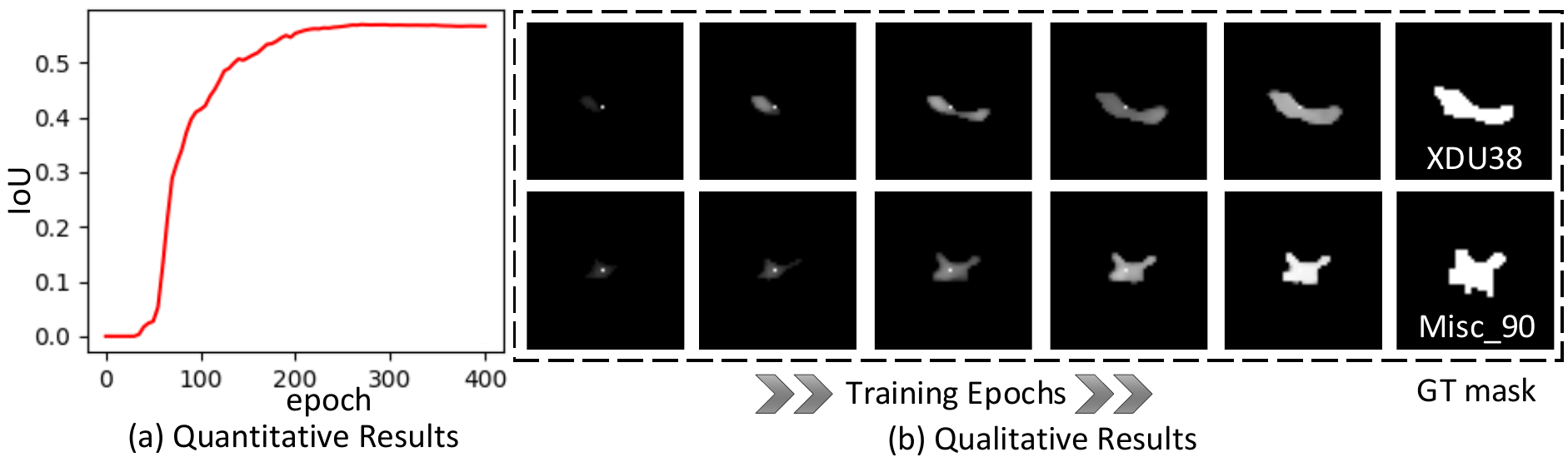}
		\vspace{-0.6cm}
		\caption{Quantitative and qualitative results of evolved target masks.}\label{fig:visual_e}
		\vspace{-0.4cm}
	\end{figure}
	
	\textbf{Effectiveness.}
	We compare the average results of {SIRST\cite{ALCNet}}, {NUDT-SIRST} \cite{DNA-Net}, and {IRSTD-1K} \cite{ISNet} datasets achieved by DNA-Net with (\textit{i.e.,} w/) and without (\textit{i.e.,} w/o) LESPS under centroid and coarse point supervision, respectively. Note that, the results of DNA-Net w/o LESPS are calculated under extremely low threshold\footnote{With point supervision, the results of DNA-Net w/o LESPS calculated under threshold 0.5 are all zeros.} (\textit{i.e.,} 0.15) while those of DNA-Net w/ LESPS are calculated under the standard threshold (\textit{i.e.,} 0.5 \cite{DNA-Net,ISNet}). As shown in Table \ref{Tab:effect}, as compared to full supervision, the results of DNA-Net w/o LESPS are extremely low, which demonstrates that SIRST detection on single point supervision is a challenging task. 
	In contrast, DNA-Net w/ LESPS can achieve significant performance improvement under both point supervisions in terms of all the metrics, which approximate the performance of their fully supervised counterparts. Note that, $P_d$ results of DNA-Net w/o LESPS under coarse point supervision are over half lower than those under the centroid ones, while the results of DNA-Net w/ LESPS under these two kinds of point supervision are comparable. It demonstrates that LESPS can generalize well to manual annotation errors. 

\begin{figure}[t]
	\vspace{-0.25cm}
	\centering\includegraphics[width=8.5cm]{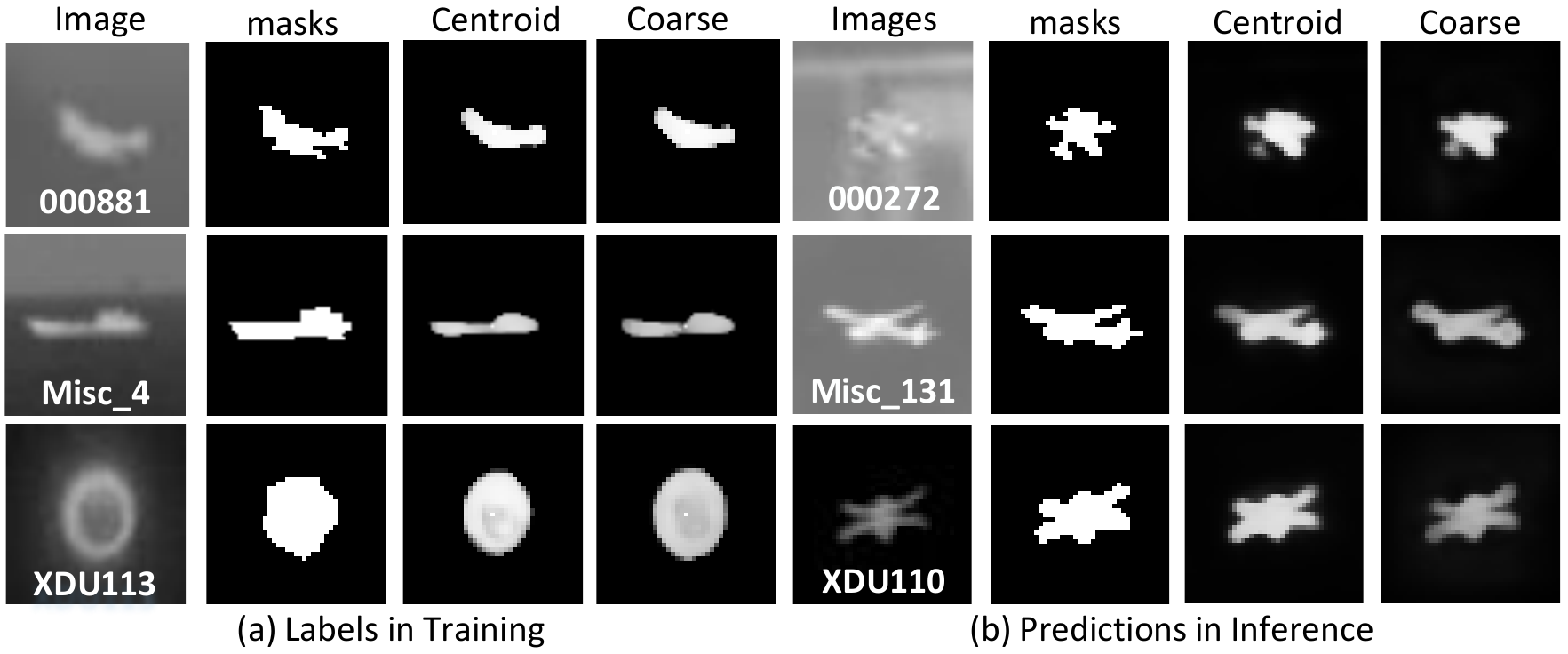}
	\vspace{-0.5cm}
	\caption{Visualizations of regressed labels during training and network predictions during inference with centroid and coarse point supervision.}\label{fig:visual_b}
	\vspace{-0.4cm}
\end{figure}

	In addition, we make evaluations of evolved target masks during training. Quantitative results in Fig.~\ref{fig:visual_e}(a) show average $IoU$ values between positive pixels of evolved target masks and GT labels in 20-time training, which demonstrates that the quality of pseudo target masks consecutively increases during training. Qualitative results in Fig.~\ref{fig:visual_e}(b) demonstrate that networks can effectively learn to expand point labels to mask labels. 
	Furthermore, we visualize the labels regressed by our LESPS during training together with some network predictions during inference in Figs.~\ref{fig:visual_b} (a) and (b). As shown in Fig.~\ref{fig:visual_b}(a), compared with GT mask labels, the evolved labels are more closely aligned with the objects in images (\textit{e.g.,} GT masks of Misc\_4, XDU113 exceed the target regions due to visual edge ambiguity), which demonstrates that LESPS can alleviate the manually annotated errors. Figure~\ref{fig:visual_b}(b) shows that DNA-Net w/ LESPS can effectively achieve accurate pixel-level SIRST detection in an end-to-end manner. Please refer to the supplemental materials for more visual results. 

	\begin{table*}
		\centering
		\scriptsize
		\vspace{-0.2cm}
			\caption{$IoU$ ($\times 10^2$), $P_d$ ($\times 10^2$) and $F_a$($\times 10^6$) values of different methods achieved on {SIRST\cite{ALCNet}}, {NUDT-SIRST} \cite{DNA-Net} and {IRSTD-1K} \cite{ISNet} datasets. ``CNN Full'', ``CNN Centroid'', and ``CNN Coarse'' represent CNN-based methods under full supervision, centroid and coarse point supervision. ``+" represents CNN-based methods equipped with LESPS.}\label{tab-performance}
			\setlength{\tabcolsep}{2.4mm}
			{\vspace{-0.2cm}
				\begin{tabular}{|l|l|ccc|ccc|ccc|ccc|}
					\hline
					\multirow{2}*{Methods}&\multirow{2}*{Description}&\multicolumn{3}{c|}{SIRST\cite{ALCNet}}&\multicolumn{3}{c|}{NUDT-SIRST \cite{DNA-Net} } &\multicolumn{3}{c|}{IRSTD-1K \cite{ISNet}} &\multicolumn{3}{c|}{Average}\\\cline{3-14}
					&&$IoU$&$P_d$&$F_a$&$IoU$&$P_d$&$F_a$&$IoU$&$P_d$&$F_a$&$IoU$&$P_d$&$F_a$\\\hline
					Top-Hat\cite{tophat}&Filtering&7.14&79.84&1012.00&20.72&78.41&166.70&10.06&75.11&1432.00&12.64&77.79&870.23  \\
					RLCM\cite{RLCM}&Local Contrast& 21.02&80.61&199.15&15.14&66.35&163.00&14.62&65.66&17.95&16.06&68.70&98.77 \\
					TLLCM \cite{TLLCM}&Local Contrast&11.03&79.47&7.27&7.06&62.01&46.12&5.36&63.97&4.93&7.22&65.45&21.42 \\
					MSPCM\cite{MSPCM}&Local Contrast& 12.38&83.27&17.77&5.86&55.87&115.96&7.33&60.27&15.24&7.23&61.53&55.13 \\
					IPI  \cite{IPI }&Low Rank& 25.67&85.55&11.47&17.76&74.49&41.23&27.92&81.37&16.18&23.78&80.47&22.96 \\
					PSTNN \cite{PSTNN}&Low Rank&22.40&77.95&29.11&14.85&66.13&44.17&24.57&71.99&35.26&20.61&72.02&36.18 \\\hline
					MDvsFA\cite{MDvsFA}&CNN Full& 61.77&92.40&64.90&45.38&86.03&200.71&35.40&85.86&99.22&47.52&88.10&121.61 \\
					ISNet \cite{ISNet}&CNN Full&72.04&94.68&42.46&71.27&96.93&96.84&60.61&94.28&61.28&67.97&95.30&66.86\\
					UIU-Net \cite{UIU-Net}&CNN Full&69.90&95.82&51.20&75.91&96.83&18.61&61.11&92.93&26.87&68.97&95.19&32.23 \\\hline
					\multirow{3}*{ACM\cite{ACM}} &CNN Full&64.92&90.87&12.76&57.42&91.75&39.73&57.49&91.58&43.86&59.94 &91.40 &32.12 \\
					&\cellcolor{mygray}{CNN Centroid+}&\cellcolor{mygray}{49.23}&\cellcolor{mygray}{89.35}&\cellcolor{mygray}{40.95}&\cellcolor{mygray}{42.09}&\cellcolor{mygray}{91.11}&\cellcolor{mygray}{38.24}&\cellcolor{mygray}{41.44}&\cellcolor{mygray}{88.89}&\cellcolor{mygray}{60.46}&\cellcolor{mygray}{44.25}&\cellcolor{mygray}{89.78}&\cellcolor{mygray}{46.55} \\
					&\cellcolor{mygray}{CNN Coarse+}&\cellcolor{mygray}{47.81}&\cellcolor{mygray}{88.21}&\cellcolor{mygray}{40.75}&\cellcolor{mygray}{40.64}&\cellcolor{mygray}{81.11}&\cellcolor{mygray}{49.45}&\cellcolor{mygray}{40.37}&\cellcolor{mygray}{92.59}&\cellcolor{mygray}{64.81}&\cellcolor{mygray}{42.94 }&\cellcolor{mygray}{87.30 }&\cellcolor{mygray}{51.67} \\\hline
					\multirow{3}*{ALCNet \cite{ALCNet}}&CNN Full&67.91&92.78&37.04&61.78&91.32&36.36&62.03&90.91&42.46&63.91 &91.67 &38.62 \\
					&\cellcolor{mygray}{CNN Centroid+}&\cellcolor{mygray}{50.62}&\cellcolor{mygray}{92.02}&\cellcolor{mygray}{36.84}&\cellcolor{mygray}{41.58}&\cellcolor{mygray}{92.28}&\cellcolor{mygray}{67.01}&\cellcolor{mygray}{44.90}&\cellcolor{mygray}{90.57}&\cellcolor{mygray}{84.68}&\cellcolor{mygray}{45.70}&\cellcolor{mygray}{91.62}&\cellcolor{mygray}{62.84} \\
					&\cellcolor{mygray}{CNN Coarse+}&\cellcolor{mygray}{51.00}&\cellcolor{mygray}{90.87}&\cellcolor{mygray}{42.40}&\cellcolor{mygray}{44.14}&\cellcolor{mygray}{92.80}&\cellcolor{mygray}{32.10}&\cellcolor{mygray}{46.75}&\cellcolor{mygray}{92.26}&\cellcolor{mygray}{64.30}&\cellcolor{mygray}{47.30}&\cellcolor{mygray}{91.98}&\cellcolor{mygray}{46.27} \\\hline
					\multirow{3}*{DNA-Net \cite{DNA-Net}} &CNN Full&76.86&96.96&22.5&87.42&98.31&24.5&62.73&93.27&21.81&75.67 &96.18 &22.94 \\
					&\cellcolor{mygray}{CNN Centroid+}&\cellcolor{mygray}{61.95}&\cellcolor{mygray}{92.02}&\cellcolor{mygray}{18.17}&\cellcolor{mygray}{57.99}&\cellcolor{mygray}{94.71}&\cellcolor{mygray}{26.45}&\cellcolor{mygray}{52.09}&\cellcolor{mygray}{88.88}&\cellcolor{mygray}{16.09}&\cellcolor{mygray}{57.34}&\cellcolor{mygray}{91.87}&\cellcolor{mygray}{20.24} \\
					&\cellcolor{mygray}{CNN Coarse+}&\cellcolor{mygray}{61.13}&\cellcolor{mygray}{93.16}&\cellcolor{mygray}{11.87}&\cellcolor{mygray}{58.37}&\cellcolor{mygray}{93.76}&\cellcolor{mygray}{28.01}&\cellcolor{mygray}{49.05}&\cellcolor{mygray}{87.54}&\cellcolor{mygray}{15.07}&\cellcolor{mygray}{56.18}&\cellcolor{mygray}{91.49}&\cellcolor{mygray}{18.32} \\
					\hline
			\end{tabular}}
		\vspace{-0.2cm}
	\end{table*}

	\textbf{Initial Evolution Loss.}
	We investigate the optimal value of loss-based threshold $T_{loss}$. Fig.~\ref{fig:LB} (a) shows that high initial evolution loss results in significant difference between $PA$ and $IoU$ (\textit{i.e.,} 0.94 vs. 0.04 with  $T_{loss}$=$50$). That is, over-large initial evolution loss introduces much error pixels into labels, resulting in severe error accumulation and thus leads to network convergence failure. Reducing initial evolution loss can reduce error accumulation and promote network convergence (0.60 vs. 0.54 with  $T_{loss}$=$10$). However, over-small initial evolution loss, which represents high degree of mapping degeneration, results in inferior performance (0.21 vs. 0.21 with  $T_{loss}$=$2$). Therefore, we choose $T_{loss}=10$ in our method.
	
	
	\textbf{Evolution Threshold.} \label{sec:exp-thresh}
	We investigate the optimal values of $k$ and $T_b$ in the evolution threshold. $T_b$ is the minimum threshold, and controls evolution speed and error accumulation degree. $k$ determines the maximum threshold, and controls the growth rate of the threshold. As shown in Fig.~\ref{fig:LB}(b) and \ref{fig:LB}(c), both over-large and over-small values of $T_b$ and $k$ result in inferior performance. Therefore, we choose $k$=1/2 and $T_b$=0.5 in our method.

	\textbf{Evolution Frequency.}
	We investigate the optimal value of evolution frequency $f$. Figure~\ref{fig:LB}(d) shows that evolution frequency is positively correlated to $PA$ and $IoU$. However, high evolution frequency needs more time for label updates. To balance performance and efficiency, we choose $f$=5 in our method. Interestingly, higher frequency (\textit{i.e.,} $f$=2) does not cause serve error accumulation, which also demonstrates the effectiveness of the convergence conditions of our LESPS. Please refer to the supplemental materials for more discussions of the convergence issue.
	
	\begin{figure}[t]
		\vspace{-0.3cm}
		\centering\includegraphics[width=8.5cm]{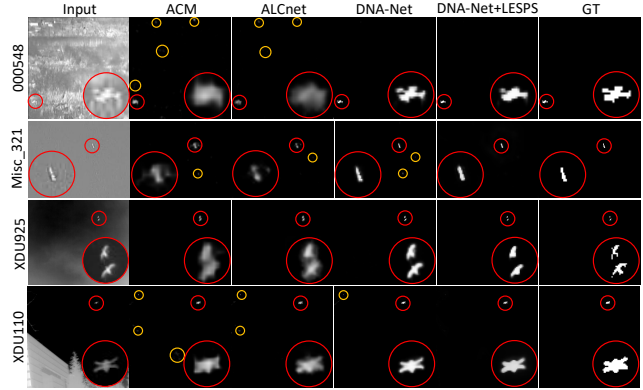}
		\caption{{Visual detection results of different methods achieved on {SIRST\cite{ALCNet}}, {NUDT-SIRST} \cite{DNA-Net} and {IRSTD-1K} \cite{ISNet} datasets. Correctly detected targets and false alarms are highlighted by red and orange circles,
				respectively.}}\label{fig:visual1}%
		\vspace{-0.4cm}
	\end{figure}
	
	\subsection{Comparison to State-of-the-art Methods}
	
	\textbf{Comparison to SISRT detection methods.}
	We apply our LESPS to several state-of-the-art CNN-based methods, including ACM \cite{ACM}, ALCNet \cite{ALCNet} and DNA-Net \cite{DNA-Net}. For fair comparisons, we retrained all models on SIRST \cite{ALCNet}, NUDT-SIRST \cite{DNA-Net}, and IRSTD-1K \cite{ISNet} datasets with the same settings. In addition, we add the results of six fully supervised CNN-based methods (MDvsFA \cite{MDvsFA}, ACM \cite{ACM}, ALCNet \cite{ALCNet}, DNA-Net \cite{DNA-Net}, ISNet \cite{ISNet}, UIU-Net \cite{UIU-Net}) and six traditional methods (Top-Hat \cite{tophat}, RLCM \cite{RLCM}, TLLCM \cite{TLLCM}, MSPCM \cite{WSLCM}, IPI \cite{IPI}, PSTNN \cite{PSTNN}) as the baseline results. 
	
	Quantitative results in Table \ref{tab-performance} show that CNN-based methods equipped with LESPS can outperform all the traditional methods. In addition, they can also achieve 71-75\% $IoU$ results and comparable $P_d$ and $F_a$ results of their fully supervised counterparts.
	Qualitative results in Fig.~\ref{fig:visual1} show that CNN-based methods equipped with LESPS can produce outputs with precise target mask and low false alarm rate, and can generalize well to complex scenes.
	Please refer to supplemental materials for more quantitative and qualitative results.
	
	\textbf{Comparison to other pseudo labels.}
	We compare our dynamic updated pseudo labels with fixed pseudo labels generated by input intensity threshold and local contrast-based methods \cite{RLCM,MSPCM,TLLCM}. Specifically, given a GT point label, we only preserve the eight connected regions of detection results that have union pixels with the GT point label. Then, we employ the pseudo labels to retrain DNA-Net \cite{DNA-Net} from scratch. As shown in Table~\ref{tab-pseudo}, compared with fixed pseudo labels, dynamic updated pseudo labels by LESPS can achieve the highest $IoU$ values with comparable $P_d$ and reasonable $F_a$ increase. 
	\begin{table}
		\centering
		\setlength{\tabcolsep}{2.5mm}
		\caption{Average $IoU$ ($\times 10^2$), $P_d$ ($\times 10^2$), $F_a$($\times 10^6$) values on {SIRST\cite{ALCNet}}, {NUDT-SIRST} \cite{DNA-Net} and {IRSTD-1K} \cite{ISNet} datasets of DNA-Net trained with pseudo labels generated by input
			intensity threshold, LCM-based methods \cite{RLCM,MSPCM,TLLCM} and LESPS under centroid and coarse point supervision.}\label{tab-pseudo} 
		\vspace{-0.2cm}
		\scriptsize
		{\begin{tabular}{|l|ccc|ccc|}
				\hline
				\multirow{2}*{Pseudo Label}&\multicolumn{3}{c|}{Centroid}&\multicolumn{3}{c|}{Coarse}\\\cline{2-7}
				&$IoU$&$P_d$&$F_a$&$IoU$&$P_d$&$F_a$\\\hline
				Threshold=0.3&4.92&81.78&13.18&5.67&83.12&11.98\\
				Threshold=0.5&13.24&73.08&5.31&15.54&76.03&4.89\\
				Threshold=0.7&14.51&45.50&4.28&15.21&46.88&3.84\\
				RLCM\cite{RLCM}&21.43&89.10&\textbf{2.67}&22.53&90.56&3.69 \\
				TLLCM \cite{TLLCM}&21.95&90.96&7.72&26.05&\textbf{94.15}&4.27 \\
				MSPCM\cite{MSPCM}&28.89&\textbf{92.62}&3.84&29.79&93.95&\textbf{2.28} \\
				\cellcolor{mygray}{LESPS}(ours)&\cellcolor{mygray}{\textbf{57.34}}&\cellcolor{mygray}{91.87}&\cellcolor{mygray}{20.24}&\cellcolor{mygray}{\textbf{56.18}}&\cellcolor{mygray}{91.49}&\cellcolor{mygray}{18.32}\\
				\hline
		\end{tabular}}
		\vspace{-0.3cm}
	\end{table}

	\subsection{Discussion of Loss Function}
	In this subsection, we investigate the loss function of computing negative loss on different background points. Average results of different baseline methods under centroid point supervision are shown in Table~\ref{background_points}. Extremely limited handcrafted background points\footnote{Points are sampled near targets, and are fixed during training.} leads to many false alarms. Random sample\footnote{Points are randomly sampled, and change in each training iteration.} introduces more background points and well alleviates class imbalance, resulting in great performance improvements. However, the above simple versions introduce huge false alarms (34-1.8K times of all points), which are not practical for real applications, but inspire further thorough investigation in the future. 
	
	\begin{table}[t]
		\centering
		\footnotesize
		\caption{Average $IoU$($\times 10^2$), $P_d$($\times 10^2$), $F_a$($\times 10^3$) values on {SIRST\cite{ALCNet}}, {NUDT-SIRST} \cite{DNA-Net} and {IRSTD-1K} \cite{ISNet} datasets of different methods when computing negative loss on $i$ handcrafted (hand$_i$), $j$ randomly sampled (rand$_j$) and all background points.}\label{background_points}
		\vspace{-0.2cm}
		\scriptsize
		\setlength{\tabcolsep}{0.9mm}
		{\begin{tabular}{|c|ccc|ccc|ccc|}
				\hline
				\multirow{2}*{Annotations}&\multicolumn{3}{c|}{ACM}&\multicolumn{3}{c|}{ALCNet}&\multicolumn{3}{c|}{DNA-Net}\\\cline{2-10}
				&$IoU$&$P_d$&$F_a$&$IoU$&$P_d$&$F_a$&$IoU$&$P_d$&$F_a$\\\hline
				hand$_1$ &0.54&95.79&47.06&0.16&95.19&262.06&1.43&97.80&26.91 \\
				hand$_2$ &0.12&\textbf{97.24}&295.17&0.15&96.62&248.05&3.41&\textbf{98.18}&8.48 \\
				hand$_5$ &0.11&96.36&316.37&0.08&\textbf{97.29}&363.25&3.68&98.13&7.29 \\
				rand$_1$ &8.06&93.45&3.56&8.57&92.97&3.21&18.74&94.69&0.58  \\
				rand$_2$ &10.78&92.72&2.22&10.78&91.16&2.71&22.85&94.81&0.42 \\
				rand$_5$ &\textbf{13.39}&92.66&1.35&\textbf{11.87}&93.26&0.89&\textbf{24.80}&95.00&0.34 \\
				\cellcolor{mygray}{All (Ours)}&\cellcolor{mygray}{3.95}&\cellcolor{mygray}{87.15}&\cellcolor{mygray}{\textbf{0.02}}&\cellcolor{mygray}{4.08}&\cellcolor{mygray}{88.93}&\cellcolor{mygray}{\textbf{0.02}}&\cellcolor{mygray}{5.12}&\cellcolor{mygray}{89.18}&\cellcolor{mygray}{\textbf{0.01}}\\\hline
		\end{tabular}}
		\vspace{-0.4cm}
	\end{table}
	
	\section{Conclusion}
	In this paper, we proposed the first work to achieve weakly-supervised SIRST detection using single-point supervision. In our method, we discovered the mapping degeneration phenomenon and proposed a label evolution framework named label evolution with single point supervision (LESPS) to automatically achieve point-to-mask regression. Through LESPS, networks can be trained to achieve SIRST detection in an end-to-end manner. 
	Extensive experiments and insightful visualizations have fully demonstrated the effectiveness and superiority of our method. In addition, our method can be applied to different networks to achieve over 70\% and 95\% of its fully supervised performance on pixel-level $IoU$ and object-level $P_d$, respectively. We hope our interesting findings and promising results can inspire researchers to rethink the feasibility of achieving state-of-the-art performance in SIRST detection with much weaker supervision. 
	
	{\small
		\bibliographystyle{ieee_fullname}
		\bibliography{egbib}
	}
	
	\clearpage
	\setcounter{section}{0}
	\setcounter{figure}{0}
	\setcounter{table}{0}
	
	\renewcommand\thesection{\Alph{section}} 
	\renewcommand\thetable{\Roman{table}}
	\renewcommand\thefigure{\Roman{figure}}
	
	\title{Mapping Degeneration Meets Label Evolution: Learning Infrared Small Target Detection with Single Point Supervision (\textit{Supplemental Material})}
	\author{}
	
	\maketitle
	
	\begin{figure}[t]
		\vspace{-1.2cm}
		\centering\includegraphics[width=8.5cm]{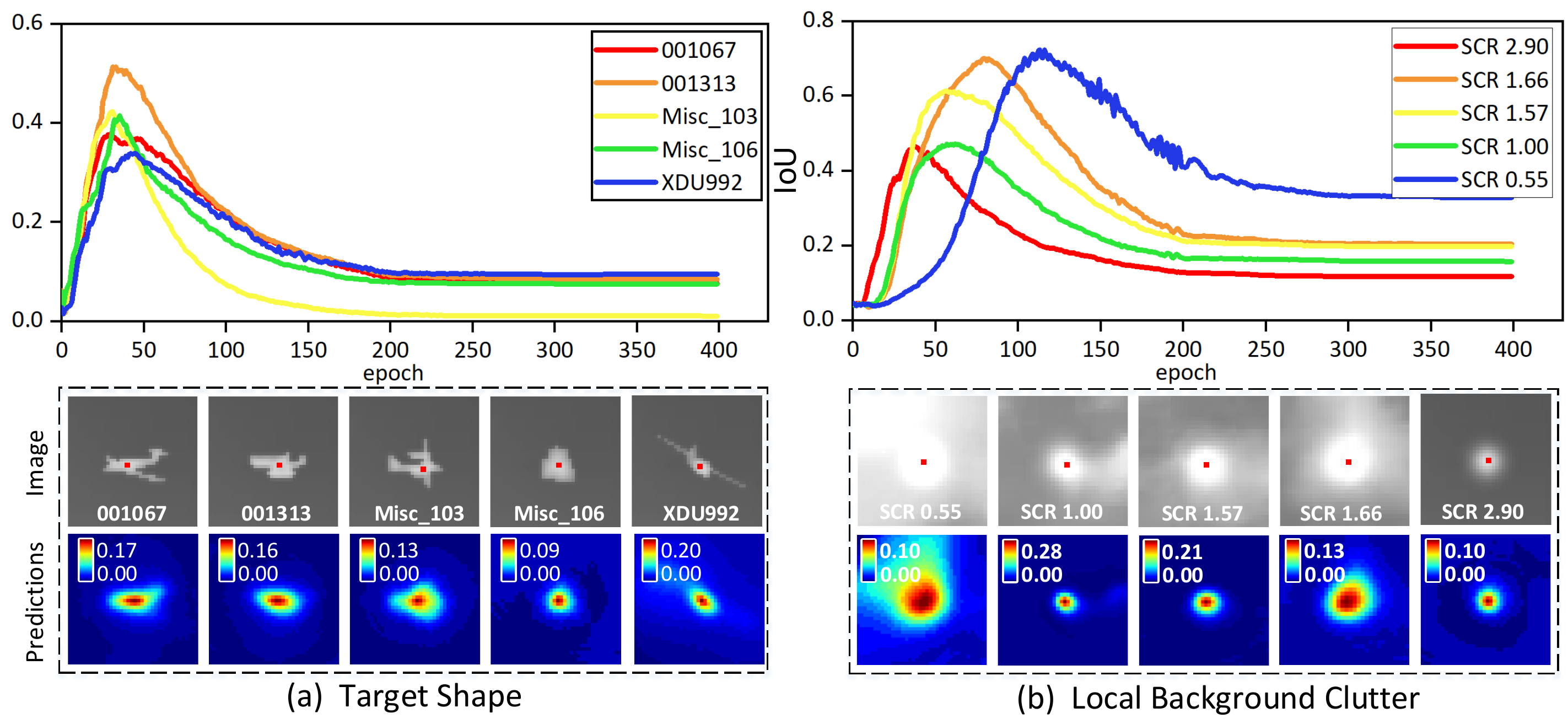}
		\caption{$IoU$ and visualize results of mapping degeneration with respect to different characteristics of targets ((a) shape and (b) local background clutter). We visualize the zoom-in target regions of input images with GT point labels (\textit{i.e.,} red dots in images) and corresponding CNN predictions (in the epoch reaching maximum $IoU$).}\label{fig:mp_Quan_supp}
		\vspace{-.5cm}
	\end{figure}
	
	Section~\ref{sec:MP} investigates the mapping degeneration phenomenon with respect to different characteristics of targets (\textit{i.e.,} shape and local background clutter) for the analyses in Section 5.2.1-\textit{Analyses of Mapping Degeneration}. Section~\ref{sec:Effectiveness} presents more visual results of labels and network predictions for the analyses in Section 5.2.2-\textit{Effectiveness}. Section~\ref{sec:Discussions} provides additional discussion of the convergence issue for the analyses in Section 5.2.2-\textit{Evolution Frequency}. 
	Section~\ref{sec:QQresults} includes additional comparison results for the analyses in Section 5.3. Section~\ref{sec:WSSM} provides comparison results with existing weakly-supervised segmentation methods. 
	
	\begin{figure}[t]
		\vspace{-1.2cm}
		\centering\includegraphics[width=8.5cm]{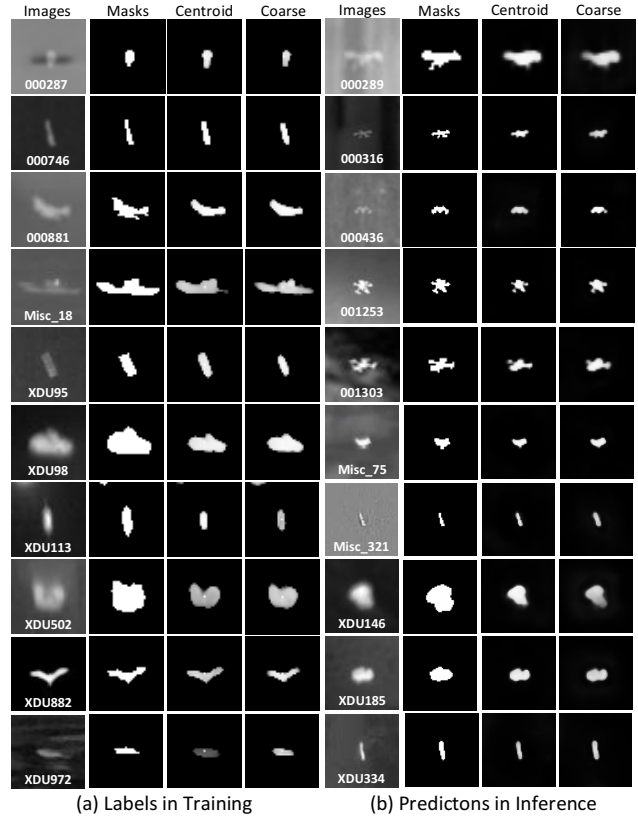}
		\caption{Visualizations of regressed labels during training and network predictions during inference with centroid and coarse point supervision.}\label{fig:visual}
		\vspace{-.5cm}
	\end{figure}

	\section{Analyses of Mapping Degeneration}
	\label{sec:MP}
	
	In this section, we investigate the mapping degeneration phenomenon with respect to different characteristics of targets (\textit{i.e.,} shape, and local background clutter).
	
	\textbf{Target Shape.} We simulate targets \cite{DNA-Net} with different shapes (\textit{i.e.,} 001067, 001313, Misc\_103, Misc\_106, XDU992) to investigate the influence of target shape on mapping degeneration. Note that, we try to keep the target size and intensity unchanged when changing the target shape. Quantitative results in Fig.~\ref{fig:mp_Quan_supp}(a) show that more concentrated shape results in higher maximum $IoU$.
	Visualization results show that CNNs can predict a cluster of pixels in a shape-aware manner, but can only recover the main body of targets without fine-grained details (\textit{e.g., }wings of drones in 001067 and XDU992).
	
	\textbf{Local Background Clutter.} We simulate a Gaussian-based extended target (with intensity 100 \& radius 7), and add them to different locations of the background image to investigate the influence of local background clutter on mapping degeneration. We employ SCR of the local neighborhood to quantify the local background clutter. Results in Fig.~\ref{fig:mp_Quan_supp}(b) show that background clutters significantly change the observed target appearance in size, shape, and contrast, and our method can only predict the high-contrast regions in the input images. Therefore, high-contrast background clutters introduce false alarms, and thus degrade the detection performance.
	
	\section{Visual Results of Labels and Network Predictions}
	\label{sec:Effectiveness}
	
	In this section, we provides additional visual results of regressed labels during training and network predictions during inference on SIRST\cite{ALCNet}, {NUDT-SIRST} \cite{DNA-Net}, and {IRSTD-1K} \cite{ISNet} datasets. It can be observed from Figure~\ref{fig:visual} that our LESPS can effectively regress mask labels during training, and can achieve accurate pixel-level SIRST detection in the inference stage.
	
	\section{Discussion of Convergence Issue}\label{sec:Discussions}
\begin{figure}[t]
		\centering\includegraphics[width=8.5cm]{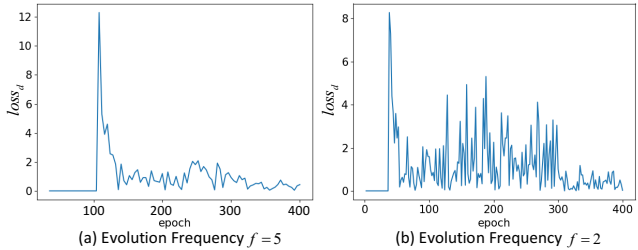}
		\caption{$loss_d$ with respect to evolution frequency $f$=$5$ and $f$=$2$.}\label{fig:loss_ba}
		\vspace{-.4cm}
	\end{figure}

	In this section, we discuss the convergence issue of our label evolution framework (\textit{i.e.,} LESPS). Specifically, we calculate the focal loss between evolved labels and network predictions before and after label update, and use their absolute difference (\textit{i.e.,} $loss_d$) to measure the proximity degree between predictions and labels. Fig.~\ref{fig:loss_ba} shows $loss_d$ of each update with evolution frequency $f$=$5$ and $f$=$2$ (\textit{i.e.,} update every 5 and 2 epochs). It can be observed that $loss_d$ is gradually reduced in general, which reveals that predictions gradually approximate labels and networks can converge steadily. In addition, the training process of $loss_d$ with $f$=$5$ is more steady than that of $f$=$2$. This is because, a relatively lower update frequency (\textit{i.e.,} $f$=$5$) represents more training time before label update, and thus stabilize the training process.
		
	\section{Quantitative and Qualitative Results}\label{sec:QQresults}
	
	\subsection{Comparison to SISRT Detection Methods}
	Table~\ref{tab-performance_supp} provides additional comparative results of six traditional methods (Max-Median \cite{Max-Median}, WSLCM \cite{WSLCM}, WSPCM \cite{WSLCM}, NRAM \cite{NRAM}, RIPT \cite{RIPT},  MSLSTIPT \cite{MSLSTIPT}). It can be observed that CNN-based methods equipped with LESPS can outperform all the traditional methods, and can achieve over 70\% $IoU$ and comparable $P_d$, $F_a$ values of their fully supervised counterparts.
	
	\begin{table}[t]
		\centering
		\renewcommand\arraystretch{1.1}
		\setlength{\tabcolsep}{2.5mm}
		\centering
		\caption{Average $IoU$ ($\times 10^2$), $P_d$ ($\times 10^2$), $F_a$($\times 10^6$) values on 3 public datasets \cite{ALCNet,DNA-Net,ISNet} of DNA-Net trained with pseudo labels generated by input
			intensity threshold, LCM-based methods \cite{RLCM,MSPCM,TLLCM} and LESPS under centroid and coarse point supervision. Best results are shown in boldface.}\label{tab-pseudo_supp}
		\vspace{-0.2cm}
		\scriptsize
		{\begin{tabular}{|l|ccc|ccc|}
				\hline
				\multirow{2}*{Pseudo Label}&\multicolumn{3}{c|}{Centroid}&\multicolumn{3}{c|}{Coarse}\\\cline{2-7}
				&$IoU$&$P_d$&$F_a$&$IoU$&$P_d$&$F_a$\\\hline
				Threshold=0.3&4.92&81.78&13.18&5.67&83.12&11.98\\
				Threshold=0.5&13.24&73.08&5.31&15.54&76.03&4.89\\
				Threshold=0.7&14.51&45.50&4.28&15.21&46.88&3.84\\\hline
				RLCM\cite{RLCM}&21.43&89.10&2.67&22.53&90.56&3.69 \\
				WSLCM\cite{WSLCM}&8.68&86.64&50.10&8.89&84.45&80.24 \\
				TLLCM \cite{TLLCM}&21.95&90.96&7.72&26.05&\textbf{94.15}&4.27 \\
				MSLCM \cite{MSLCM}&31.43&\textbf{93.16}&\textbf{2.50}&36.32&92.43&\textbf{1.17}\\
				MSPCM\cite{MSPCM}&28.89&92.62&3.84&29.79&93.95&2.28 \\\hline
				Ours&\textbf{57.34}&91.87&20.24&\textbf{56.18}&91.49&18.32 \\
				\hline
		\end{tabular}}
	\end{table}
	
	\begin{table}[t]
		\centering
		\footnotesize
		\renewcommand\arraystretch{1.1}
		\caption{Average $IoU$($\times 10^2$), $P_d$($\times 10^2$), $F_a$($\times 10^6$) values of different methods. ``\#Params." represents the number of parameters. Best results are shown in boldface.}\label{comparison}  
		\vspace{-0.2cm}
		\scriptsize
		\setlength{\tabcolsep}{0.8mm}
		{\begin{tabular}{|l|l|c|ccc|}
				\hline
				Method&Annotations per object&\#Params.&$IoU$&$P_d$&$F_a$\\\hline 
				MaskRCNN+ [C3]&10 points in bbox&88.6M&51.30&\textbf{94.38}&82.77 \\
				PointRend [C4]&100+ elaborated points&120.3M&56.02&94.30&61.48 \\ 
				Implicit PointRend [C3]&10 points in bbox&700.0M&52.00&94.13&85.79\\
				DNA-Net+LESPS (Ours)&1 coarse point&\textbf{4.8M}&\textbf{56.18}&91.49&\textbf{18.32}\\\hline 
		\end{tabular}}
		\vspace{-0.4cm}
	\end{table}
	
	\begin{table*}
		\centering
		\renewcommand\arraystretch{1.1}
		\setlength{\tabcolsep}{2.1mm}
		\caption{$IoU$ ($\times 10^2$), $P_d$ ($\times 10^2$) and $F_a$($\times 10^6$) values of different methods achieved on {SIRST\cite{ALCNet}} {NUDT-SIRST} \cite{DNA-Net} and {IRSTD-1K} \cite{ISNet} datasets. ``CNN Full'', ``CNN Centroid'', and ``CNN Coarse'' represent CNN-based methods under full supervision, centroid and coarse point supervision. ``+" represents CNN-based methods equipped with LESPS.}\label{tab-performance_supp}
		\vspace{-0.1cm}
		\scriptsize
		{\begin{tabular}{|l|l|ccc|ccc|ccc|ccc|}
				\hline
				\multirow{2}*{Methods}&\multirow{2}*{Description}&\multicolumn{3}{c|}{SIRST\cite{ALCNet}}&\multicolumn{3}{c|}{NUDT-SIRST \cite{DNA-Net} } &\multicolumn{3}{c|}{IRSTD-1K \cite{ISNet}} &\multicolumn{3}{c|}{Average}\\\cline{3-14}
				&&$IoU$&$P_d$&$F_a$&$IoU$&$P_d$&$F_a$&$IoU$&$P_d$&$F_a$&$IoU$&$P_d$&$F_a$\\\hline
				Top-Hat\cite{tophat}&Filtering&7.14&79.84&1012.00&20.72&78.41&166.70&10.06&75.11&1432.00&12.64&77.79&870.23  \\
				Max-Median\cite{Max-Median}&Filtering&4.17&69.20&55.33&4.20&58.41&36.89&7.00&65.21&59.73&5.12&64.27&50.65 \\
				RLCM\cite{RLCM}&Local Contrast& 21.02&80.61&199.15&15.14&66.35&163.00&14.62&65.66&17.95&16.06&68.70&98.77 \\
				WSLCM\cite{WSLCM} &Local Contrast&1.02&80.99&45846.16&0.85&74.60&52391.63&0.99&70.03&15027.08&0.91&74.82&33759.07 \\
				TLLCM \cite{TLLCM}&Local Contrast&11.03&79.47&7.27&7.06&62.01&46.12&5.36&63.97&4.93&7.22&65.45&21.42 \\
				MSLCM \cite{MSLCM}&Local Contrast& 11.56&78.33&8.37&6.65&56.83&25.62&5.35&59.93&5.41&7.07&61.20&13.74 \\
				MSPCM\cite{MSPCM}&Local Contrast& 12.38&83.27&17.77&5.86&55.87&115.96&7.33&60.27&15.24&7.23&61.53&55.13 \\
				IPI  \cite{IPI }&Low Rank& 25.67&85.55&11.47&17.76&74.49&41.23&27.92&81.37&16.18&23.78&80.47&22.96 \\
				NRAM \cite{NRAM}&Low Rank&12.16&74.52&13.85&6.93&56.40&19.27&15.25&70.68&16.93&11.45&67.20&16.68  \\
				RIPT \cite{RIPT}& Low Rank&11.05&79.08&22.61&29.44&91.85&344.30&14.11&77.55&28.31&18.20&82.83&131.74 \\
				PSTNN \cite{PSTNN}&Low Rank&22.40&77.95&29.11&14.85&66.13&44.17&24.57&71.99&35.26&20.61&72.02&36.18 \\
				MSLSTIPT \cite{MSLSTIPT}&Low Rank&10.30&82.13&1131.00&8.34&47.40&888.10&11.43&79.03&1524.00&10.02&69.52&1181.03 \\\hline
				MDvsFA\cite{MDvsFA}&CNN Full& 61.77&92.40&64.90&45.38&86.03&200.71&35.40&85.86&99.22&47.52&88.10&121.61 \\
				ISNet \cite{ISNet}&CNN Full&72.04&94.68&42.46&71.27&96.93&96.84&60.61&94.28&61.28&67.97&95.30&66.86\\
				UIU-Net \cite{UIU-Net}&CNN Full&69.90&95.82&51.20&75.91&96.83&18.61&61.11&92.93&26.87&68.97&95.19&32.23 \\\hline
				\multirow{3}*{ACM\cite{ACM}} &CNN Full&64.92&90.87&12.76&57.42&91.75&39.733&57.49&91.58&43.86&59.94 &91.40 &32.12 \\
				&\cellcolor{mygray}{CNN Centroid+}&\cellcolor{mygray}{49.23}&\cellcolor{mygray}{89.35}&\cellcolor{mygray}{40.95}&\cellcolor{mygray}{42.09}&\cellcolor{mygray}{91.11}&\cellcolor{mygray}{38.24}&\cellcolor{mygray}{41.44}&\cellcolor{mygray}{88.89}&\cellcolor{mygray}{60.46}&\cellcolor{mygray}{44.25}&\cellcolor{mygray}{89.78}&\cellcolor{mygray}{46.55} \\
				&\cellcolor{mygray}{CNN Coarse+}&\cellcolor{mygray}{47.81}&\cellcolor{mygray}{88.21}&\cellcolor{mygray}{40.75}&\cellcolor{mygray}{40.64}&\cellcolor{mygray}{81.11}&\cellcolor{mygray}{49.45}&\cellcolor{mygray}{40.37}&\cellcolor{mygray}{92.59}&\cellcolor{mygray}{64.81}&\cellcolor{mygray}{42.94 }&\cellcolor{mygray}{87.30 }&\cellcolor{mygray}{51.67} \\\hline
				\multirow{3}*{ALCNet \cite{ALCNet}}&CNN Full&67.91&92.78&37.04&61.78&91.32&36.36&62.03&90.91&42.46&63.91 &91.67 &38.62 \\
				&\cellcolor{mygray}{CNN Centroid+}&\cellcolor{mygray}{50.62}&\cellcolor{mygray}{92.02}&\cellcolor{mygray}{36.84}&\cellcolor{mygray}{41.58}&\cellcolor{mygray}{92.28}&\cellcolor{mygray}{67.01}&\cellcolor{mygray}{44.90}&\cellcolor{mygray}{90.57}&\cellcolor{mygray}{84.68}&\cellcolor{mygray}{45.70}&\cellcolor{mygray}{91.62}&\cellcolor{mygray}{62.84} \\
				&\cellcolor{mygray}{CNN Coarse+}&\cellcolor{mygray}{51.00}&\cellcolor{mygray}{90.87}&\cellcolor{mygray}{42.40}&\cellcolor{mygray}{44.14}&\cellcolor{mygray}{92.80}&\cellcolor{mygray}{32.10}&\cellcolor{mygray}{46.75}&\cellcolor{mygray}{92.26}&\cellcolor{mygray}{64.30}&\cellcolor{mygray}{47.30}&\cellcolor{mygray}{91.98}&\cellcolor{mygray}{46.27} \\\hline
				\multirow{3}*{DNA-Net \cite{DNA-Net}} &CNN Full&76.86&96.96&22.5&87.42&98.31&24.5&62.73&93.27&21.81&75.67 &96.18 &22.94 \\
				&\cellcolor{mygray}{CNN Centroid+}&\cellcolor{mygray}{61.95}&\cellcolor{mygray}{92.02}&\cellcolor{mygray}{18.17}&\cellcolor{mygray}{57.99}&\cellcolor{mygray}{94.71}&\cellcolor{mygray}{26.45}&\cellcolor{mygray}{52.09}&\cellcolor{mygray}{88.88}&\cellcolor{mygray}{16.09}&\cellcolor{mygray}{57.34}&\cellcolor{mygray}{91.87}&\cellcolor{mygray}{20.24} \\
				&\cellcolor{mygray}{CNN Coarse+}&\cellcolor{mygray}{61.13}&\cellcolor{mygray}{93.16}&\cellcolor{mygray}{11.87}&\cellcolor{mygray}{58.37}&\cellcolor{mygray}{93.76}&\cellcolor{mygray}{28.01}&\cellcolor{mygray}{49.05}&\cellcolor{mygray}{87.54}&\cellcolor{mygray}{15.07}&\cellcolor{mygray}{56.18}&\cellcolor{mygray}{91.49}&\cellcolor{mygray}{18.32} \\
				\hline
		\end{tabular}}
	\end{table*}
	
	\begin{figure*}[t]
		\centering
		\includegraphics[width=17.5cm]{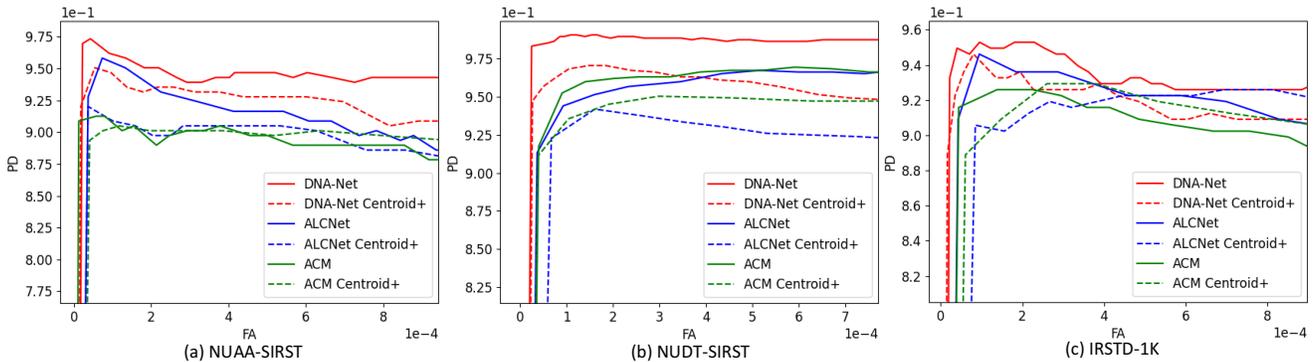}
		\caption{ROC results of different methods achieved on (a) SIRST\cite{ALCNet}, (b) {NUDT-SIRST} \cite{DNA-Net}, and (c) {IRSTD-1K} \cite{ISNet} datasets. ``Centroid+'' represents CNN-based methods equipped with LESPS under centroid point supervision.}\label{fig:ROC}
	\end{figure*}
	
	\begin{figure*}[t]
		\centering
		\includegraphics[width=17.5cm]{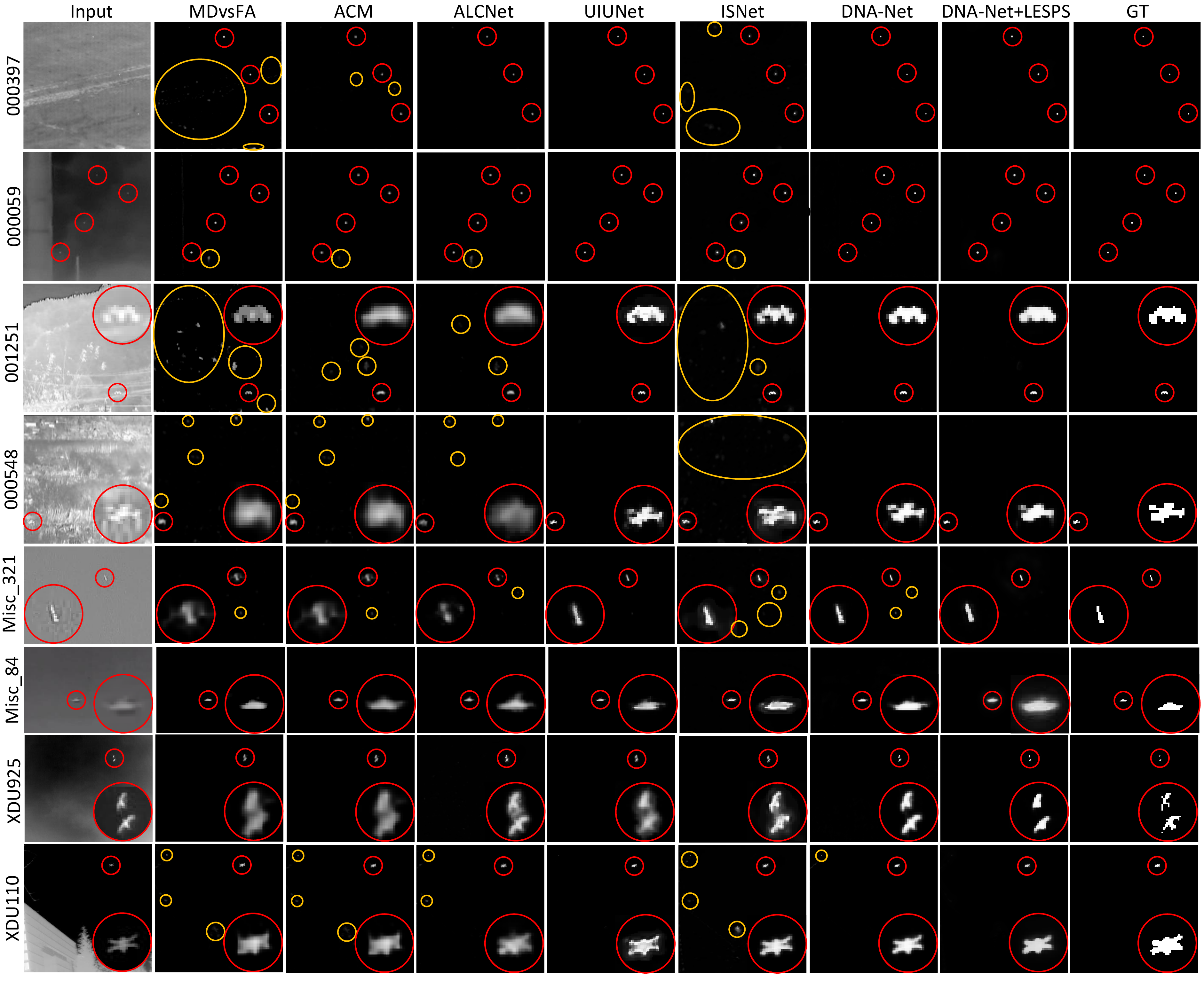}
		\caption{Visual detection results of different methods achieved on {SIRST\cite{ALCNet}}, {NUDT-SIRST} \cite{DNA-Net}, and {IRSTD-1K} \cite{ISNet} datasets. Correctly detected targets and false alarms are highlighted by red and orange circles, respectively.}\label{fig:visual2}
	\end{figure*}                                                                                                                                                                                                                                 
	
	Figure~\ref{fig:ROC} provides ROC results of ACM \cite{ACM}, ALCNet \cite{ALCNet}, DNA-Net \cite{DNA-Net} equipped with LESPS under centroid point supervision (\textit{i.e.,} ACM Centroid+, ALCNet Centroid+, DNA-Net Centroid+) and their fully supervised counterparts (\textit{i.e.,} ACM, ALCNet, DNA-Net) achieved on SIRST \cite{ALCNet}, {NUDT-SIRST} \cite{DNA-Net}, and {IRSTD-1K} \cite{ISNet} datasets. It can be observed that ROC results of ACM Centroid+, ALCNet Centroid+, DNA-Net Centroid+, and ACM, ALCNet, DNA-Net only have minor differences (\textit{i.e.,} less than 5\%).
	
	Figure~\ref{fig:visual2} provides additional qualitative results. It can be observed that CNN-based methods equipped with LESPS can produce outputs with precise target mask and low false alarm rate, and can generalize well to complex scenes.
	
	\subsection{Comparison to Fixed Pseudo Labels}
	Table~\ref{tab-pseudo_supp} provides additional comparisons to more LCM-based pseudo labels. It can be observed that, compared with LCM-based pseudo labels, DNA-Net with LESPS can achieve the highest IoU values with comparable $P_d$ and reasonable $F_a$ increase.
	
	\section{Comparison to Existing Weakly-Supervised Segmentation Methods}\label{sec:WSSM}
	We equip DNA-Net with LESPS, and compare with existing weakly-supervised segmentation methods\footnote{All models are implemented by their officially public codes.}. Results are shown in Table~\ref{comparison}. It can be observed that general weakly-supervised segmentation methods require much more annotation effort and computational cost (\textit{i.e.,} 18-146 times of our method) but the performance is comparable or worse. It is demonstrated that different from general methods, point-supervised SISRT detection has its unique characteristics, and needs further exploration.

	{\small
		\bibliographystyle{ieee_fullname}
		\bibliography{egbib}
	}

\end{document}